\documentclass[11pt,a4paper]{article}
\usepackage[hyperref]{eacl2021}
\usepackage{times}
\usepackage{latexsym}
\usepackage{graphicx}
\usepackage{multirow}
\usepackage{tcolorbox}

\usepackage{paralist}

  \pltopsep=\medskipamount
\usepackage{microtype}

\aclfinalcopy 

\title{Towards objectively evaluating \\ the quality of generated medical summaries}

\author{
    Francesco Moramarco\thanks{~ Equal contribution} \\
  Babylon Health / London, UK \\
  University of Aberdeen / Aberdeen, UK \\
  \texttt{francesco.moramarco\textsuperscript{†}} \\\And
  Damir Juric\textsuperscript{*} \\
  Babylon Health / London, UK \\
  \texttt{damir.juric\textsuperscript{†}} \\
  \AND
  Aleksandar Savkov \\
  Babylon Health / London, UK \\
  \texttt{sasho.savkov\textsuperscript{†}} \\\And
  Ehud Reiter \\
  University of Aberdeen / Aberdeen, UK \\
  \texttt{e.reiter@abdn.ac.uk} \\
  \AND
  \textsuperscript{†}\normalfont{\texttt{@babylonhealth.com}}
}

\date{}

\begin{document}
\maketitle

\begin{abstract}

We propose a method for evaluating the quality of generated text by asking evaluators to count facts, and computing precision, recall, f-score, and accuracy from the raw counts. We believe this approach leads to a more objective and easier to reproduce evaluation. We apply this to the task of medical report summarisation, where measuring objective quality and accuracy is of paramount importance.



\end{abstract}

\section{Introduction}
\label{sec:intro}

Natural Language Generation in the medical domain is notoriously hard because of the sensitivity of the content and the potential harm of hallucinations and inaccurate statements \cite{kryscinski2020evaluating, falke2019ranking}. This informs the human evaluation of NLG systems, selecting accuracy and overall quality of the generated text as the most valuable aspects to be evaluated.

In this paper we carry out a human evaluation of the quality of medical summaries of Clinical Reports generated by state of the art (SOTA) text summarisation models.

Our contributions are: (i) a re-purposed parallel dataset of medical reports and summary descriptions for training and evaluating, (ii) an approach for a more objective human evaluation using counts, and (iii) a human evaluation conducted on this dataset using the approach proposed.
\section{Related Work}
\label{sec:related}


A recent study by \citet{celikyilmaz2020evaluation} gives a comprehensive view on different approaches to text summary evaluation. While many of these can be wholly or partly translated between different domains, the medical domain remains particularly problematic due to the sensitive nature of its data.
\citet{moen2014evaluation} and \citet{MOEN201625} try to establish if there is a correlation between automatic and human evaluations of clinical summaries. A 4-point and 2-point Likert scales are used for the human evaluation.
In \citet{goldstein2017evaluation} the authors generate free-text summary letters from the data of 31 different patients and compare them to the respective original physician-composed discharge letters, measuring relative completeness, quantifying missed data items, readability, and functional performance.

Closest to our approach is the Pyramid method by \citet{nenkova2007}, which defines semantically motivated, sub-sentential units (Summary Content Units) for annotators to extract in each reference summary. SCUs are weighed according to how often they appear in the multiple references and then compared with the SCUs extracted in the hypothesis to compute precision, recall, and f-score.


\section{Data}
\label{sec:data}

The \emph{MTSamples} dataset comprises 5,000 sample medical transcription reports from a wide variety of specialities uploaded to a community platform website\footnote{\url{https://mtsamples.com}}. The dataset has been used in past medical NLP research \cite{Chen2011,Lewis2011,Soysal2017} including as a Kaggle dataset\footnote{\url{https://www.kaggle.com/tboyle10/medicaltranscriptions}}.

There are $40$ medical specialties in the dataset, such as `Surgery', `Consult - History and Phy.', and `Cardiovascular / Pulmonary'. Each specialty contains a number of sample reports ranging from $6$ to $1103$. 

\begin{figure}[h]
    \centering
    \includegraphics[width=0.48\textwidth]{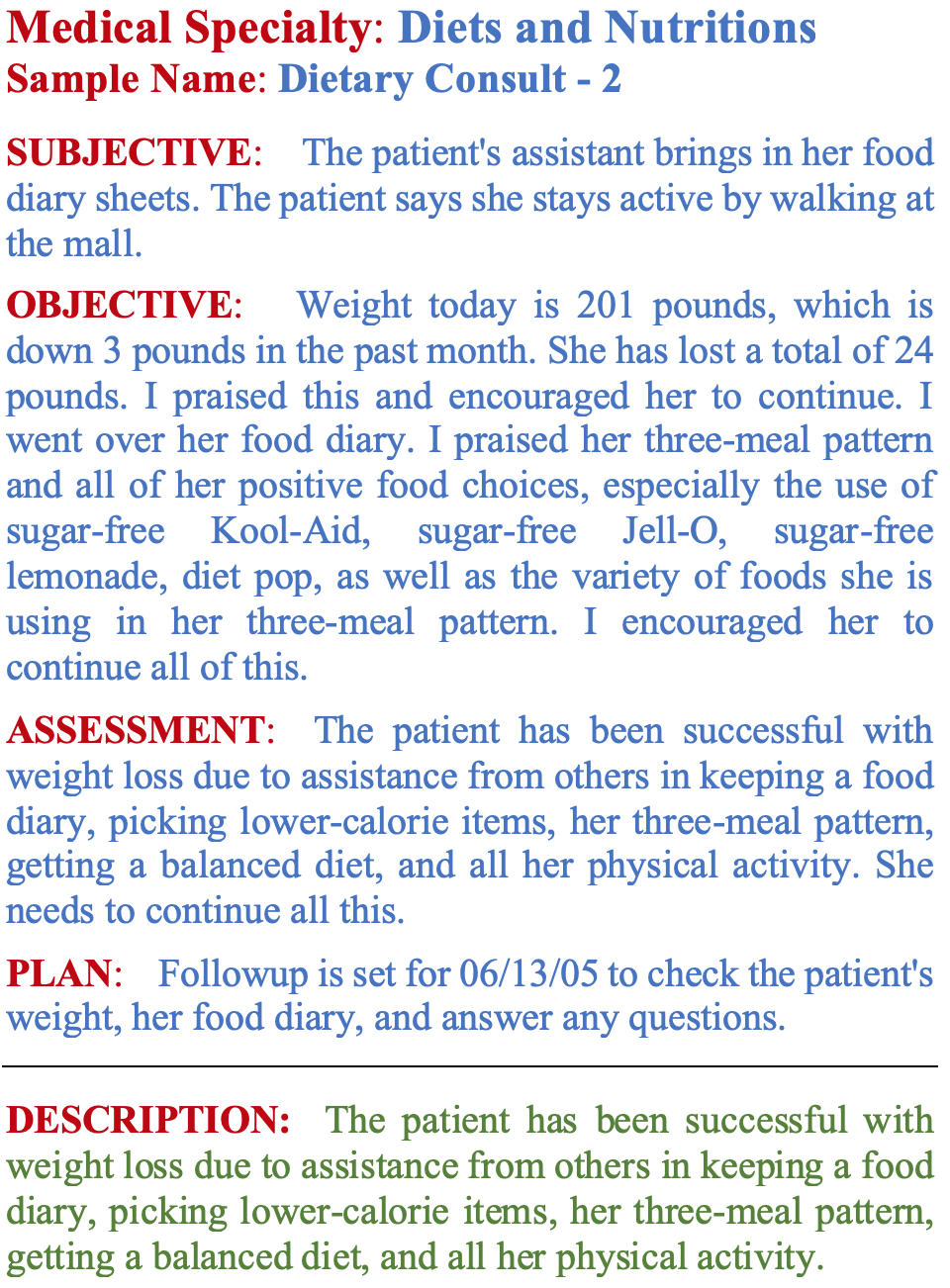}
    \caption{An MTSamples clinical report of specialty `Diets and Nutritions'. Note the reference Description at the bottom.}
    \label{fig:clinical-report-example}
\end{figure}

The reports are free text with headings, which change according to the specialty. However, all reports also have a description field, which is a good approximation of a summary of the report. The length of each report varies greatly according to the specialty, with an average of $589$ words for the body of the report, and $21$ words for the description. Figure \ref{fig:clinical-report-example} shows an example of MTSamples reports, inclusive of description.

Given the brevity of some descriptions, we discard reports with descriptions shorter than $12$ words and consider a dataset of $3242$ reports.
By examining the dataset, we note that descriptions are mostly extractive in nature, meaning they are phrases or entire sentences taken from the report. To quantify this we compute n-gram overlap with Rouge-1 (unigram) and Rouge-L (longest common n-gram) \cite{lin2004rouge} precision scores, which are $0.989$ and $0.939$ respectively.

We split the dataset into $2{\,}576$ reports for training ($80\%$), $323$ for development ($10\%$) and $343$ for testing ($10\%$). We perform the split separately for each medical specialty to ensure they are adequately represented and then aggregate the data.

The dataset, models, and evaluation results can be found on Github\footnote{\url{https://github.com/babylonhealth/medical-note-summarisation}}.



\section{Experimental Setup}
\label{sec:experimental_setup}

For our experiment, we consider one baseline and three SOTA automatic summarisation models (extractive, abstractive, and fine-tuned on our training set respectively). More specifically:

\vspace{-0.05in}
\begin{itemize}
    \item \textbf{Lead–3} — this is our baseline. Following \citet{zhang2018neural}, this model selects the first three sentences of the clinical report as the description;
    \item \textbf{Bert–Ext} — the unsupervised extractive model by \citet{miller2019leveraging} \footnote{\url{https://pypi.org/project/bert-extractive-summarizer/}};
    \item \textbf{Pegasus–CNN} — an abstractive model by \citet{zhang2019pegasus} trained on the CNN/Daily mail dataset and used as is;
    \item \textbf{Bart–Med} — an abstractive model by \citet{lewis-etal-2020-bart}, which we fine-tune on our MTSamples training set. 
\end{itemize}

We generate descriptions with these $4$ models using the entire clinical report text as input.
\section{Human Evaluation Protocol}
\label{sec:human_evaluation}

We select $10$ clinical reports and summary descriptions from our MTSamples test set. Our subjects are three general practice physicians. They are employed at Babylon Health and have experience in AI research evaluation.
The task is implemented with the Heartex Annotation Platform\footnote{\url{https://www.heartex.ai/}}, which lets researchers define tasks in an XML language and specify the number of annotators. It then generates each individual task and collates the results.

The task involves (i) reading the clinical report, (ii) reading the reference description (supplied by the dataset, see Figure \ref{fig:clinical-report-example}), (iii) then evaluating $4$ generated descriptions by answering 5 questions (for a total of 40 generated descriptions). We ask the evaluators to count the ``medical facts'' in each generated description and to compare them against those in the reference.
Initially, we considered listing the types of facts to be extracted, as done by \citet{thomson2020gold}, but the sheer diversity in the structure and content across the specialties in our dataset made this approach impractical. Instead, we give evaluators instructions containing two examples and ask them to extrapolate a process for fact extraction. Figure \ref{fig:instructions} shows the instructions we give them.

\begin{figure}[tbh]
\begin{tcolorbox}[boxrule=2pt,arc=.3em,boxsep=1mm]
 The evaluation consists of reading a clinical report and a number of short descriptions, then quantifying how many ``medical facts'' were correctly reported. We understand that the definition of a ``medical fact'' is vague, and so it's up to your interpretation. As an example, in the following description:

\begingroup
\small
\begin{verbatim}
2-year-old female who comes in for
just rechecking her weight, her
breathing status, and her diet.
\end{verbatim}
\endgroup

    There are (arguably) 4 facts:
    \vspace{-0.06in}
    \begin{itemize}
        \item 2 year old female
        \vspace{-0.04in}
        \item coming to recheck her weight
        \vspace{-0.04in}
        \item coming to recheck her breathing status
        \vspace{-0.22in}
        \item coming to recheck her diet
    \end{itemize}

Here's another example:

\begingroup
\small
\begin{verbatim}
The patient had a syncopal episode 
last night. She did not have any 
residual deficit. She had a headache
at that time. She denies chest pains
or palpitations.
\end{verbatim} 
\endgroup

Here there are (arguably) 5 facts:
\vspace{-0.06in}
\begin{itemize}    
    \item had a syncopal episode last night
    \vspace{-0.04in}
    \item no residual deficit
    \vspace{-0.04in}
    \item headache
    \vspace{-0.04in}
    \item no chest pains
    \vspace{-0.04in}
    \item no palpitations
\end{itemize}

\end{tcolorbox}

\caption{Instructions to evaluators.}
\label{fig:instructions}
\end{figure}

\begin{figure}[tbh]
    \centering
    \includegraphics[width=0.5\textwidth]{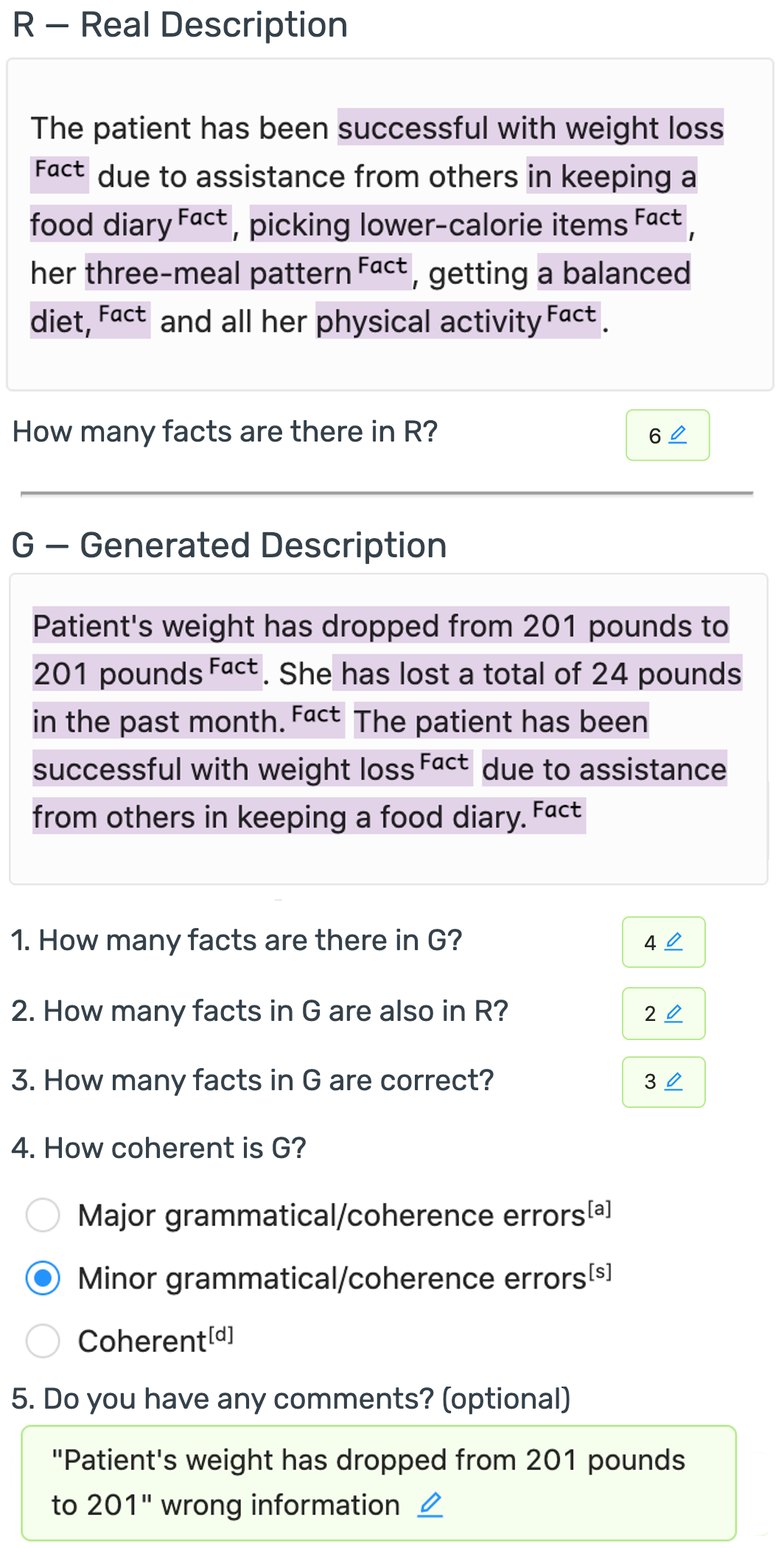}
    \caption{A completed task. \emph{Real Description} is the reference.}
    \label{fig:annotated-task}
\end{figure}

The evaluators are asked to read the clinical report (as shown in Figure \ref{fig:clinical-report-example}), then to analyse the reference description by reporting the number of facts. To aid them in the task, they can optionally select the facts in the text using an in-built Heartex feature. Next, they are shown four generated descriptions (one per model) and asked to count facts and answer 5 questions. Figure \ref{fig:annotated-task} shows the reference, generated descriptions, and questions for a given task, and gives an example annotation from one of the evaluators.
When answering question 3 (How many facts in G are correct?) they refer to the clinical report as a ground truth.

Based on this set of questions, we gather the following raw counts:
\vspace{-0.06in}

\begin{itemize}
    \item $R$: facts in the reference description
    \item $G$: facts in the generated description
    \item $R\&G$: facts in common
    \item $C$: correct facts in the generated description
\end{itemize}

We use these raw counts to compute four derived metrics:
\vspace{-0.06in}

\begin{itemize}
    \item \textbf{Precision}, calculated as $\frac{R\&G}{G}$ 
    \item \textbf{Recall}, calculated as $\frac{R\&G}{R}$
    \item \textbf{F-Score}, calculated as $2 \cdot \frac{Precision \cdot Recall}{Precision + Recall}$
    \item \textbf{Accuracy}, calculated as $\frac{C}{G}$
\end{itemize}

For \emph{Coherence}, we take \citet{chen-etal-2020-shot} and \citet{juraska-etal-2019-viggo} definition: ``\emph{whether the generated text is grammatically correct and fluent, regardless of factual correctness}'' and ask evaluators to choose between three options (Coherent, Minor Errors, Major Errors) and convert these into continuous numbers with Coherent = 1.0, Minor Errors = 0.5, and Major Errors = 0.0.
\section{Results and Discussion}
\label{sec:results}

\begin{table}[t]
    \centering
    \setlength\tabcolsep{2pt} 
    \begin{tabular}{c|c|ccc|c}
        \textbf{Model} & \textbf{Metric} & \textbf{Eval 1} & \textbf{Eval 2} & \textbf{Eval 3} & \textbf{Avg} \\
        \hline
        \hline
        \multirow{5}{*}{\rotatebox[origin=c]{90}{Lead-3}}
        & Precision & 0.42 & 0.43 & 0.46 & 0.44 \\
        & Recall    & 0.64 & 0.60 & 0.73 & 0.66\\
        & F-Score   & 0.49 & 0.45 & 0.51 & 0.48\\
        & Accuracy  & 1.0 & 1.0 & 1.0 & \textbf{1.0} \\
        & Coherence & 0.95 & 0.95 & 0.90 & 0.93 \\
        
        \hline
        \multirow{5}{*}{\rotatebox[origin=c]{90}{Bert-Ext} }
        & Precision & 0.58 & 0.48 & 0.48 & 0.51\\
        & Recall    & 0.62 & 0.61 & 0.60 & 0.61 \\
        & F-Score   & 0.59 & 0.52 & 0.51 & 0.54\\
        & Accuracy  & 1.0 & 1.0 & 1.0 & \textbf{1.0} \\
        & Coherence & 1.0 & 0.95 & 1.0 & \textbf{0.98}\\

        \hline
        \multirow{5}{*}{\rotatebox[origin=c]{90}{Pegasus-CNN}} 
        & Precision & 0.29 & 0.36 & 0.31 & 0.32\\
        & Recall    & 0.43 & 0.50 & 0.50 & 0.47\\
        & F-Score   & 0.34 & 0.40 & 0.36 & 0.37\\
        & Accuracy  & 0.97 & 0.97 & 0.98 & 0.97\\
        & Coherence & 1.0 & 1.0 & 0.95 & \textbf{0.98} \\

        \hline
        \multirow{5}{*}{\rotatebox[origin=c]{90}{Bart-Med}}
        & Precision & 0.65 & 0.58 & 0.55 & \textbf{0.59}\\
        & Recall    & 1.0 & 0.96 & 0.97 & \textbf{0.98}\\
        & F-score   & 0.77 & 0.70 & 0.68 & \textbf{0.72}\\
        & Accuracy  & 1.0 & 1.0 & 1.0 & \textbf{1.0}\\
        & Coherence & 1.0 & 0.75 & 0.95 & 0.90 \\

        \hline
    \end{tabular}
    \caption{Derived metrics for each model and each evaluator, aggregated across tasks.}
    \label{tab:results}
\end{table}

Table \ref{tab:results} shows the results for all derived metrics, calculated on the raw counts from the evaluators. Expectedly, Bart-Med, the model trained on the MTSamples training set, scores highest in all metrics (except Coherence).

Interestingly, all four models score almost-perfect accuracy, meaning they don't hallucinate medical facts. This is not a surprise for Lead-3 and Bert-Ext, which are extractive in nature. As for Pegasus-CNN and Bart-Med, while the models are abstractive, we notice they tend to mostly select and copy phrases or entire sentences from the source report.
The only hallucination the evaluators found is a numerical error, reported by Pegasus-CNN in the following generated description:

\begingroup
\small
\begin{verbatim}
Patient's weight has dropped from 201
pounds to 201 pounds. She has lost a
total of 24 pounds in the past month.
\end{verbatim}
\endgroup

Whereas, the source report states:

\begingroup
\small
\begin{verbatim}
Weight today is 201 pounds, which is
down 3 pounds in the past month. She
has lost a total of 24 pounds.
\end{verbatim}
\endgroup

\subsection{Agreement}
To validate the human evaluation task, we compute inter-annotator agreement for each derived metric, as well as on the raw counts. We use Krippendorff Alpha \cite{hayes2007answering} as we are dealing with continuous values. Table \ref{tab:agreement} includes overall agreement and a breakdown for each pair of evaluators.

\begin{table}[t]
    \centering
    \setlength\tabcolsep{2pt} 
    \begin{tabular}{c|c|c|ccc}
       & \textbf{Metric} & \textbf{E1–E2–E3} & \textbf{E1–E2} & \textbf{E1–E3} & \textbf{E2–E3} \\
        \hline
        \hline
        \multirow{5}{*}{\rotatebox[origin=c]{90}{Raw Counts}} & R facts & 0.25 & 0.44 & 0.27 & 0.01 \\
        & G facts & 0.33 & 0.50 & 0.26 & 0.12 \\
        & G\&R facts & 0.55 & 0.74 & 0.50 & 0.40 \\
        & G acc facts & 0.34 & 0.51 & 0.27 & 0.13 \\
        & Coherence & 0.40 & 0.14 & 0.56 & 0.49 \\
        \hline
        \multirow{4}{*}{\rotatebox[origin=c]{90}{Der. Metrics}} & Precision & 0.87 & 0.84 & 0.88 & 0.88\\
        & Recall & 0.90 & 0.93 & 0.89 & 0.89 \\
        & F-Score & 0.89 & 0.88 & 0.91 & 0.87 \\
        & Accuracy & 0.87 & 0.79 & 0.96 & 0.84 \\
    \end{tabular}
    \caption{Krippendorff Alpha for each metric, where \emph{R} is reference, \emph{G} the generated description, \emph{G acc facts} the count of accurate facts in the generated description, \emph{E1-E2-E3} the agreement of all three evaluators, and \emph{Ex-Ey} the agreement between Evaluator x and Evaluator y.}
    \label{tab:agreement}
\end{table}

Looking at the \emph{E1-E2-E3} column, we note a clear divide between the low agreement on raw counts and the high agreement on the derived metrics. We investigate this by comparing the facts selected by each annotator and notice a degree of variability in the level of granularity they employed. Consider the description:

\begingroup
\small
\begin{verbatim}
An 83-year-old diabetic female presents
today stating that she would like dia-
betic foot care.
\end{verbatim}
\endgroup

Table \ref{tab:facts} shows the facts selected by the three evaluators.

\begin{table}[tbh]
    \centering
    \setlength\tabcolsep{2pt} 
    \begin{tabular}{c|c|l}
        \textbf{E} & \textbf{Count} & \textbf{Selected Facts} \\ \hline
        E1 & 2 & \shortstack[l]{- 83-year-old diabetic female\\- would like diabetic foot care} \\ \hline
        E2 & 5 & \shortstack[l]{- 83-year-old \\- diabetic
    \\- female \\- presents today \\- would like diabetic foot care} \\ \hline
        E3 & 3 & \shortstack[l]{- 83-year-old female \\- diabetic \\- would like diabetic foot care} \\
    \end{tabular}
    \caption{Example of evaluators disagreement in fact selection.}
    \label{tab:facts}
\end{table}

We compute pairwise agreement in Table \ref{tab:agreement} and notice that two of the evaluators (E1 and E2) share a similar (more granular) approach to fact selection, whereas E3 is less granular.

        

We also investigate the low agreement for Coherence and discover that it's due to a strong imbalance of the three classes (Coherent, Minor Errors, and Major Errors) where Coherent appears $91.67\%$ of cases, Minor Errors $6.67\%$ and Major Errors $1.67\%$. While this causes a low Krippendorff Alpha, we count the number of times all three evaluators agree on a generated description being Coherent and find it to be $82.5\%$.

Finally, for all derived metrics the agreement scores are very high. This shows a robustness of these metrics even with different granularity in fact selection, and that the three evaluators agree on the quality of a given generated description. In other words, the evaluators agree on the quality of the generated descriptions even though they don't agree on the way of selecting medical facts.


    


\section{Future Work}
\label{sec:future_work}

In this paper we presented an evaluation of the quality of medical summaries using fact counting. The results of this study help us to identify a number of insights to guide our future work:
\vspace{-0.05in}

\begin{itemize}
    \item We could work on better defining a medical fact (as in \citet{duvsek2020evaluating}) and to prompt agreement on the level of granularity, for instance by instructing evaluators to split a description into the highest number of facts that are meaningful;
    \item Our evaluation focused on the quality of the generated descriptions and did not evaluate their usefulness in the medical setting. Such extrinsic evaluation would be very valuable;

    \item We could compare our approach of fact counting with the more common Likert scales.
\end{itemize}




\bibliography{anthology,eacl2021}
\bibliographystyle{acl_natbib}

\newpage
\onecolumn
\appendix
\setcounter{table}{0}
\renewcommand{\thetable}{A\arabic{table}}
\setcounter{figure}{0}
\renewcommand{\thefigure}{A\arabic{figure}}



\end{document}